\documentclass[10pt,twocolumn,letterpaper]{article}

\usepackage{iccv}
\usepackage{times}
\usepackage{epsfig}
\usepackage{graphicx}
\usepackage{amsmath}
\usepackage{amssymb}
\usepackage{multirow}
\usepackage{gensymb}

\usepackage[breaklinks=true,bookmarks=false]{hyperref}

\iccvfinalcopy 

\def\iccvPaperID{5863} 

\ificcvfinal\fi

\begin{document}

\title{Fast Point R-CNN}

\author{Yilun Chen$^{1}$ \quad Shu Liu$^{2}$ \quad Xiaoyong Shen$^{2}$ \quad Jiaya Jia$^{1,2}$ \\
$^1$The Chinese University of Hong Kong \quad 
$^2$Tencent YouTu Lab \\
{\tt\small \{ylchen, leojia\}@cse.cuhk.edu.hk, \{shawnshuliu, dylanshen\}@tencent.com}
}

\maketitle
\thispagestyle{empty}

\begin{abstract}
We present a unified, efficient and effective framework for point-cloud based 3D object detection. Our two-stage approach utilizes both voxel representation and raw point cloud data to exploit respective advantages. The first stage network, with voxel representation as input, only consists of light convolutional operations, producing a small number of high-quality initial predictions. Coordinate and indexed convolutional feature of each point in initial prediction are effectively fused with the attention mechanism, preserving both accurate localization and context information. The second stage works on interior points with their fused feature for further refining the prediction. Our method is evaluated on KITTI dataset, in terms of both 3D and Bird's Eye View (BEV) detection, and achieves state-of-the-arts with a 15FPS detection rate. 
\end{abstract}

\section{Introduction}
One challenging task in 3D perception is 3D object detection, which serves as the basic component for perception in autonomous driving, robotics, \etc. Deep convolutional neural networks (CNN) greatly improve performance of 3D object detection~\cite{MV3D, VoxelNet, fpointnet, AVOD, PIXOR, FAF}. Recent approaches of 3D object detection utilize different types of data, including monocular~\cite{monocular3d} images, stereo images~\cite{3dproposal} and RGB-D images~\cite{slidingshape, deepsliding}. In autonomous driving, point clouds captured by LiDAR are the more general and informative data format to help make prediction \cite{MV3D, fpointnet, AVOD, FAF}.

\vspace{-0.1in}
\paragraph{Challenges} LiDAR point cloud is an essential type of geometry data for 3D detection. High sparseness and irregularity of point cloud, however, make it not easily tractable for CNN. One scheme is to transform the sparse point cloud to the volumetric representation in compact shape by discretization, which is called voxelization. This representation enables CNN to perform recognition. 

However, volumetric representation is still computationally challenging. One line of solutions is to use a coarse grid~\cite{VoxelNet, PIXOR, FAF, birdnet, complexyolo, yolo3d}; but coarse quantization prevents following CNN from utilizing fine-grained information. Several consecutive convolutional layers and subsampling operations in the CNN worsen the problem.
 
Another line~\cite{pointnet,pointnet++,pointcnn,dgcnn} is to process point cloud directly for 3D object recognition. Different from the volumetric representation, coordinates of point cloud and their structure are directly fed into the neural network to exploit precise localization information. We note applying these methods to large-scale point clouds for autonomous driving is still computationally very heavy.

\vspace{-0.1in}
\paragraph{Our Contributions} In this paper, we propose a unified, fast and effective two-stage 3D object detection framework, making use of both voxel representation and raw point cloud data. The first stage of our network, named VoxelRPN, directly exploits the voxel representation of point clouds. Computationally economical convolutional layers are adopted for both high efficiency and surprisingly high-quality detection. 

In the second stage, we apply a light-weight PointNet to further refine the predictions. With a small number of initial predictions, the second stage is also in a very fast speed. We design the module with attention mechanism to effectively fuse the coordinates of each interior point with the convolution feature from the first stage. It makes each point aware of its context information. 

One characteristic of our approach is that it benefits from both representation of point clouds in volumetric representation and raw dense coordinates. The 3D volumetric representation provides a robust way to process point clouds. The light-weight PointNet in the second stage inspects coordinates of points again to capture more localization information with enlarged receptive fields, producing decent results. Since our method utilizes convolutional feature for each region on point clouds and is with high efficiency, we name it Fast Point R-CNN.

With this conceptually simple structure, we achieve high efficiency and meanwhile decent 3D detection accuracy, achieving state-of-the-art results. It is even more effective than prior methods that take both RGB and point clouds as input. The main contribution of this paper is threefold.

\begin{itemize}
    \item We propose a quick and practical two-stage 3D object detection framework based on point clouds (without RGB images), exploiting both volumetric representation and raw dense input of point clouds.\vspace{-0.09in}
    \item Our system consists of both 2D and 3D convolution to preserve information. We fuse convolutional features with point coordinate for box refinement.\vspace{-0.09in} 
    \item Our system runs at 15FPS and achieves state-of-the-art performance in terms of BEV and 3D detection, especially for high quality object detection.
\end{itemize}

\section{Related Work}
We briefly review recent work on 3D data representation of point clouds and 3D object detection.

\vspace{-0.1in}
\paragraph{3D Data Representation}
Representation of point clouds from 3D LiDAR scanners is fundamental for different tasks. Generally there are two main ways -- voxelization~\cite{voxnet, 3dshapenet} or raw point clouds~\cite{pointnet, pointnet++, pointcnn, dgcnn}. For the first type, Maturana \etal.~\cite{voxnet} first applied 3D convolution for 3D object recognition. For the point-based approaches, PointNet~\cite{pointnet} is the pioneer to directly learn feature representation based on raw points. It further aggregates global descriptors for classification. Recently, Rethage \etal.~\cite{fcnforpointcloud} employed PointNet as the low-level feature descriptor in each 3D grid and applied 3D convolution. There are also other methods that do not process 3D data directly. For example, most view-based methods \cite{multiviewcnn, volumetricandmultiview, multiviewfromsingleimage} care more about 2D color and gather information from different views of rendered images. 

\vspace{-0.1in}
\paragraph{3D Object Detection}
Over past a few years, a series of 3D detectors \cite{MV3D, fpointnet, AVOD, VoxelNet, PIXOR, FAF, contfuse, pointrcnn, ipod, pointpillar} achieved promising results on KITTI benchmark \cite{kitti}. 

\vspace{0.05in}
\noindent{\it Joint Image-LiDAR Detection}:
Several approaches~\cite{MV3D, AVOD, contfuse, fpointnet} fused information from different sensors, such as RGB images and LiDAR. For example, MV3D~\cite{MV3D} fused BEV and front view of LiDAR points as well as images, and designed a deep fusion scheme to combine region-wise features from multiple views. AVOD~\cite{AVOD} fused BEV and images in full resolutions to improve prediction quality, especially for small objects. Accurate geometric information may be lost in the high-level layers with this scheme. Contfuse~\cite{contfuse} compensated the geometric information via combining the convolution feature over LiDAR point cloud with the nearest image features and LiDAR point coordinates in the multi-scale scheme. In spite of geometric information encoded in each voxel, deeper layers have access mostly to coarse geometric feature. Based on a strong 2D detector on image, F-PointNet~\cite{fpointnet} and PointFusion~\cite{pointfusion} incorporated PointNet structures to estimate the amodal 3D box. But the 2D detector and PointNet are two separate stages and the final results heavily rely on the 2D detection results.

\vspace{0.05in}
\noindent{\it LiDAR-based Detection}:
Most LiDAR-based detection approaches process point clouds as voxel-input and apply either 2D convolution or 3D convolution to make prediction. Due to directly encoding coordinates of point clouds into voxel grid, deep layers may gradually lose this level of information. Several encoding techniques \cite{slidingshape, deepsliding, 3dfcn, velofcn} provide other representations to preserve more information. 
Chen \etal.~\cite{MV3D} encoded hand-crafted features for respective representation of BEV and front view. Instead of hand-crafted features, VoxelNet~\cite{VoxelNet} applied VFE layers via a PointNet-like network to learn low-level geometric feature, by which it shows good performance. However the network structure is computationally heavy. Recently, SECOND~\cite{second} applied Sparse Convolution~\cite{sparseconv} to speed up VoxelNet and produce better results. PointPillars~\cite{pointpillar} applied acceleration techniques, including NVIDIA TensorRT, to achieve high speed. We note they may also accelerate our method. PointRCNN~\cite{pointrcnn} and IPOD~\cite{ipod}, concurrent with our work, generate point-wise proposals on Point Clouds, which consumes much computation on point-wise calculation in the similar region or background region.

\section{Our Method}
\begin{figure*}
    \begin{center}
        \includegraphics[width=140mm]{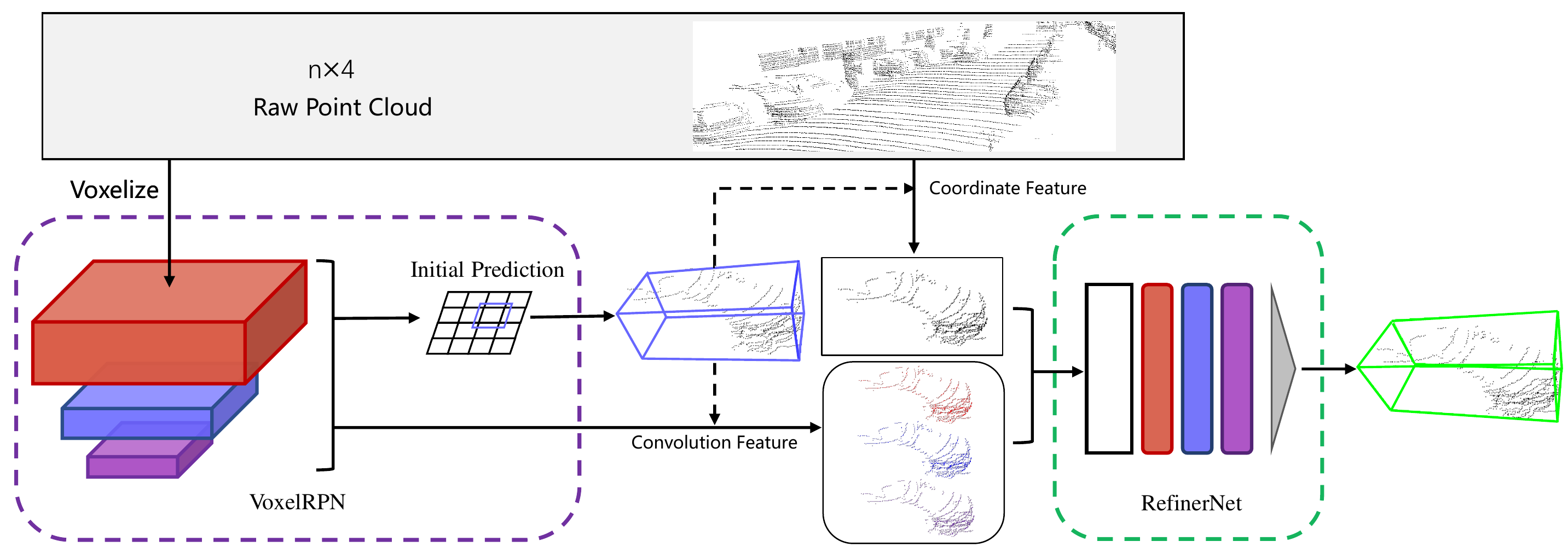}\vspace{-0.1in}
    \end{center}
    \caption{Overview of our two-stage framework. In the first stage, we voxelize point cloud and feed them to VoxelRPN to produce a small number of initial predictions. Then we generate the box feature for each prediction by fusing interior points' coordinates and context feature from VoxelRPN. Box features are fed to RefinerNet for further refinement. }
    \label{fig:overall network}
\end{figure*}

In this paper, we propose a simple and fast two-stage framework for 3D object detection with point cloud data, as shown in Figure~\ref{fig:overall network}. The first stage takes voxel representation as input and produces a set of initial predictions. To compensate the loss of precise localization information in the voxelization and consecutive convolution process, the second stage combines raw point cloud with context feature from the first stage to produce refinement results. 

\subsection{Motivation}
Point cloud, captured by LiDAR, is a set of points with irregular structure and sparse distribution. It is not straightforward to make use of powerful CNN for training and inference on point cloud data. Discretizing points into voxelized input~\cite{VoxelNet, contfuse} or projecting them to BEV with compact shape like RGB images~\cite{PIXOR, FAF} forms a set of solutions, where abstract and rich feature representation can be produced. However, the discretization process inevitably introduces quantization artifacts with resolution decreasing to the number of bins in the voxel map. Moreover, consecutive convolution and downsampling operation may also weaken the precise localization signal that originally exists in point clouds.

Methods like PointNet~\cite{pointnet} are specially designed for directly processing point cloud data. Directly applying these methods to entire point cloud, which is with a large scale in scenarios of autonomous driving, may produce more position-informative results. But they require a huge amount of GPU memory and computation, almost impossible to achieve a high detection speed. Other methods~\cite{fpointnet} rely on detection results from 2D detector followed by regression of the 3D amodal box for each object. This kind of pipeline heavily relies on 2D detection results, inheriting the weakness when detecting cluttered or distant objects in images. Clearly, directly working on point cloud data is a better choice if information can be properly made use of.

To this end, our method is new to exploit the hybrid of voxel and raw point cloud, without relying on RGB images. The two effective stages are voxel representation input to VoxelRPN to acquire a set of initial predictions in high speed, and RefinerNet to fuse raw point cloud and extracted context feature for better localization quality. These two components are elaborated on in the following.

\subsection{VoxelRPN}
VoxelRPN takes 3D voxel input and produces 3D detection results. It is a one-stage object detector.

\vspace{-0.1in}
\paragraph{Input Representation}
Input to VoxelRPN is the voxelized point cloud, which is actually a regular grid. Each voxel in the grid contains information of original points lying in the local region. Specifically, we divide the 3D space into spatially arranged voxels. Suppose the region of interest for the point cloud is a cuboid of size $(L, W, H)$ and each voxel is of size $(v_l, v_w, v_h)$, the 3D space can be divided into 3D voxel grid of size $(L / v_l, W / v_w, V / v_h)$. 

There may be more than one points in a voxel. In VoxelNet \cite{VoxelNet}, $35$ points are kept and  fed to the VFE layers to extract features. Our finding, however, is that simply using 6 points in each voxel followed a 8-channel MLP layer is already {\it adequate} to achieve
reasonable performance empirically. With this representation in a compact shape, we easily exploit the great power of CNN for informative feature extraction.

\vspace{-0.1in}
\paragraph{Network Structure}
Aiming at 3D detection, our network needs to clearly filter information from $(X,Y,Z)$ dimensions. In \cite{PIXOR, FAF}, the $Z$ dimension is simply transformed into the channels when generating the voxel representation. Then several 2D convolutions are applied. In this way, the information along $Z$ dimension vanishes quickly. As a result, detection only on BEV becomes achievable. Differently, VoxelNet~\cite{VoxelNet} keeps three separate dimensions when producing voxels followed by three 3D convolutions. It is noticed that the efficiency is decreased.

Along a more appropriate direction, we find that a number of consecutive 3D convolutions are quite effective on preserving the 3D structure. Based on this observation, our backbone network is composed of 2D and 3D convolutions, achieving high efficiency as PIXOR \cite{PIXOR} and even higher performance than VoxelNet \cite{VoxelNet}.

\begin{figure}
    \begin{center}
        \includegraphics[width=0.8\linewidth]{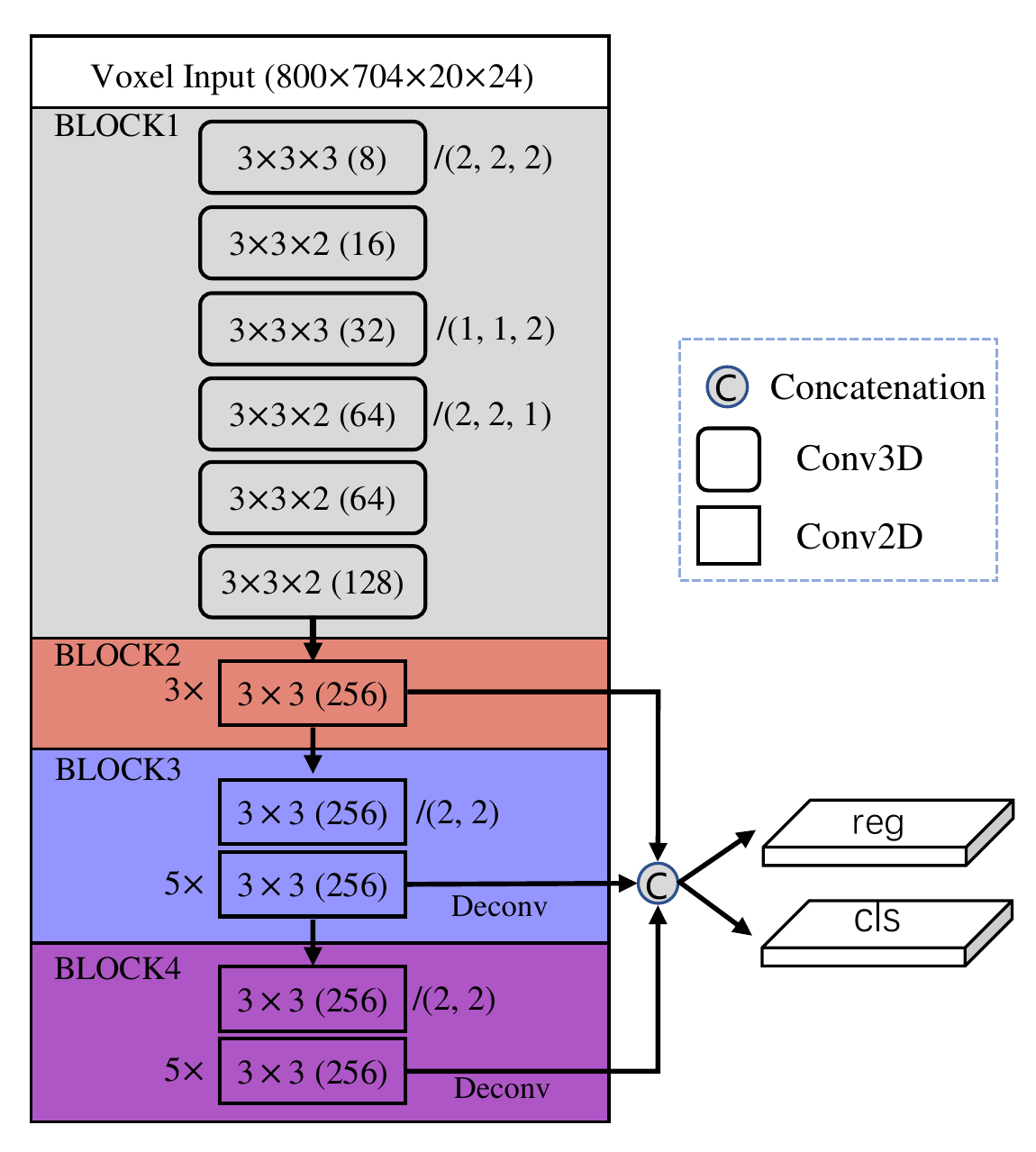}\vspace{-0.1in}
    \end{center}
    \caption{Network Structure of VoxelRPN. The format of layers used in the figure follows $(\mbox{kernel size}) (\mbox{channels})$ / $(\mbox{stride})$, \ie,  $(k_x, k_y, k_z) (\mbox{chn}) / (\text{s}_x, \text{s}_y, \text{s}_z)$. The default stride is 1 unless otherwise specified.}
    \label{fig:voxelrpn}
\end{figure}

We show details of our backbone network in Figure~\ref{fig:voxelrpn}. The first part consists of six 3D convolutional layers, which only possess a small number of filters to keep time budget. Instead of aggressively downsampling features in the $Z$ dimension by filters with stride $2$ and kernel size $3$, we insert 3D convolution layers with kernel size 2 in the $Z$ dimension without padding, to better fuse and preserve information. What follows are three blocks of 2D convolutions for further abstraction and enlarging the receptive field. 

Objects of the same category in 3D scene are generally with similar scales. Thus, different from the popular multi-scale object detector~\cite{fpn} in 2D images, which assigns object proposals to different layers according to their respective scales, we note that the HyperNet~\cite{hypernet} structure is more appropriate. 

Specifically, we upsample by deconvolution the feature maps from the last layers of the block 2, 3 and 4, as illustrated in Figure~\ref{fig:voxelrpn}. Then we concatenate them to gather rich location information in lower layers and with stronger semantic information in higher layers. Pre-defined anchors \cite{ssd} are used with specific scales and angles on this fused feature map. Then the classification and regression heads run on this feature map respectively to classify each anchor and regress the location of existing objects.

\subsection{RefinerNet}
Although decent performance is achieved by VoxelRPN, We further improve the prediction quality through directly processing raw point cloud since the voxelization process and consecutively strided convolutions in the first block still lose an amount of localization information, which however can be supplemented by further feature enhancement in our RefinerNet. 

RefinerNet makes use of the coordinates of point clouds. F-PointNet~\cite{fpointnet} is the pioneer work to utilize PointNet to regress 3D amodal bounding boxes from 2D detection results. Only interior points are used for inference without aware of context information. Our method, contrarily, also benefits from important context information.

\vspace{-0.in}
\paragraph{Box Feature} We use points in each bounding box prediction of VoxelRPN to generate box feature. Different from the two independent networks used in \cite{fpointnet}, we take not only coordinates but also features extracted from VoxelRPN as input. Convolutional feature maps from VoxelRPN capture local geometric structure of objects and gradually gather them in a hierarchical way, leading to a much larger receptive field to profit prediction. Then PointNet is applied to map each point to high-dimensional space and fuse point representation through max-pooling operation to gather information among all points with its context. 

For each predicted bounding box from VoxelRPN, we first project it to BEV. Then all points around the region of BEV box 
are used as input, as illustrated in Figure~\ref{fig:overall network}. For each point $p$ with coordinate $(x_p, y_p)$ and feature map $F$ with size $(L_F, W_F, C_F)$, we define the corresponding feature as the feature vector with $C_F$ channels at location $(\lfloor \frac{x_pL_F}{L} \rfloor, \lfloor \frac{y_pW_F}{W} \rfloor)$. We grasp the final concatenation feature map from VoxelRPN with more comprehensive information.

\begin{figure}
    \begin{center}
        \includegraphics[width=1.1\linewidth]{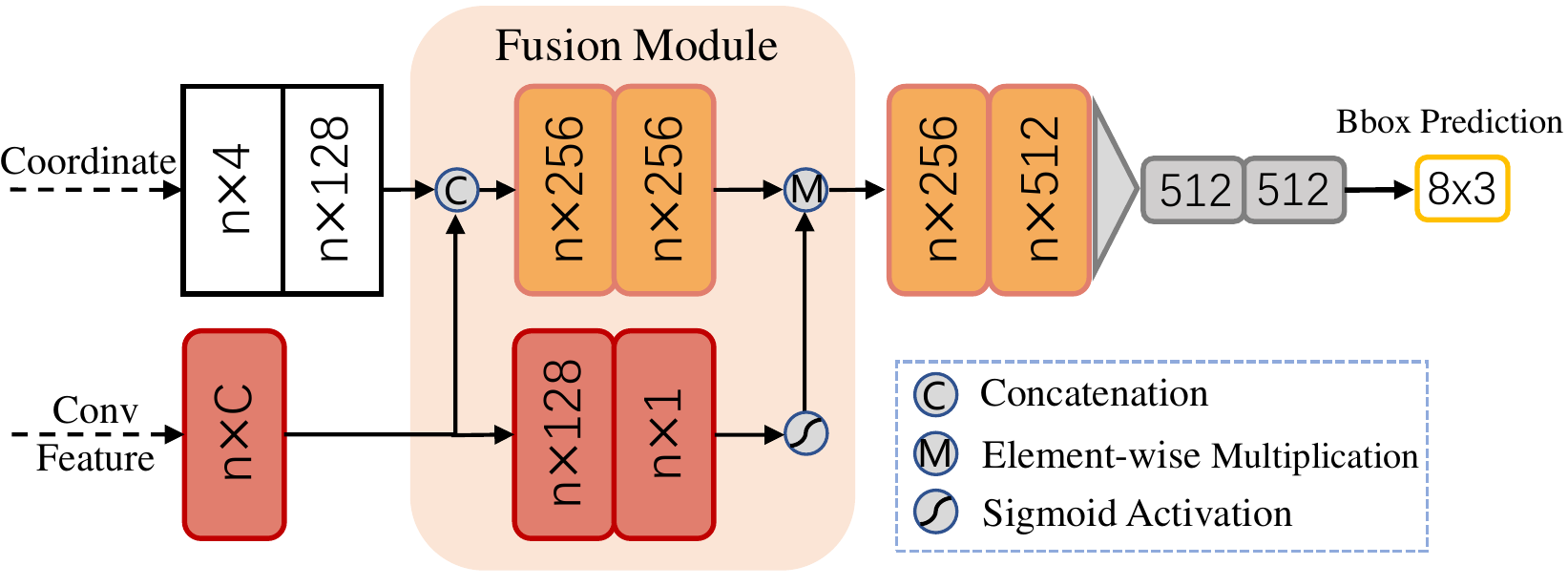}\vspace{-0.1in}
    \end{center}
    \caption{Network Structure of RefinerNet.}
    \label{fig:pointrcnn}
\end{figure}

Before feeding the coordinates of each point to the following network, we first canonize them for the purpose of guaranteeing the translation and rotation invariance. The coordinates of points within 0.3 meters around the proposal box are cropped and canonized by rotation and translation given the proposal box. As shown in Figure~\ref{fig:pointrcnn}, we define the coordinate feature as the high-dimensional (128D) representation acquired via a MLP layer. 

\vspace{-0.1in}
\paragraph{Network Structure}
With these two sources of features, we find a way to effectively fuse them. Instead of trivial concatenation, we design a new module with the attention mechanism for comprehensive feature generation. As illustrated in Figure~\ref{fig:pointrcnn}, we first concatenate the high-dimensional coordinate feature with the convolutional feature. Then it is multiplied with the attention, generated by the convolutional features. What follows is a light-weight PointNet consisting of two MLP layers with max-pooling to aggregate all information in one box. 

\begin{figure}
    \begin{center}
        \includegraphics[width=0.8\linewidth]{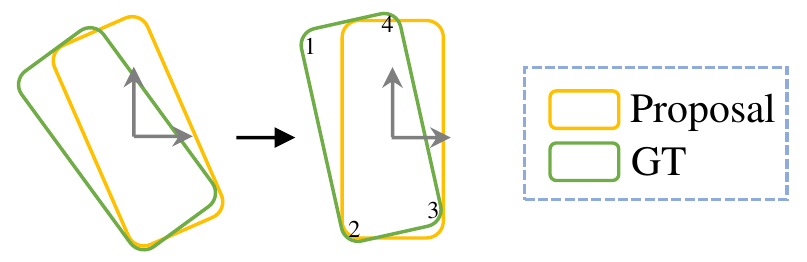}\vspace{-0.1in}
    \end{center}
    \caption{Canonization of a box. The number denotes the order of corner prediction in RefinerNet.}
    \label{fig:proposal}
\end{figure}

The final box refinement is achieved by two MLP layers to predict refined location of all box corner points based on proposals. As shown in Figure~\ref{fig:proposal}, when computing the regression target, the ground-truth box as well as point cloud are canonized by rotation and translation given the proposal box. This operation organizes ground-truth box corners in a specific order, which can reduce the uncertainty of the corner order caused by rotation. Our experiments manifest superiority of the canonized corner loss.
%

Without bells and whistles, this light-weight RefinerNet can already effectively improve the accuracy in box prediction, especially considering the $Z$ dimension and bounding boxes with higher IoUs in both 3D and BEV.

\subsection{Network Training}
Training our Fast Point R-CNN includes two steps. We first train VoxelRPN until convergence. Then the RefinerNet is trained based on the extracted features and inferred bounding boxes. 

\vspace{-0.1in}
\paragraph{VoxelRPN} In VoxelRPN, the anchors spread on each location of the global feature map. One anchor is considered as a positive sample if its IoU with ground-truth is higher than 0.6 in BEV. The regression target is the ground-truth bounding box with the highest IoU value. One anchor is considered as negative if its IoU value with all ground-truth boxes is lower than 0.45. We train VoxelRPN with a multi-task loss as 
\begin{equation}
Loss = L_{cls} +  L_{reg},
\end{equation}
where $L_{cls}$ is the classification binary cross entropy loss as
\begin{equation}
L_{cls} = \frac{1}{N_{pos}} \sum_{i} L_{cls}(p^{pos}_i, 1) + \frac{ \gamma}{N_{neg}} \sum_{i} L_{cls}(p^{neg}_i, 0),
\end{equation}
\begin{equation}
\begin{split}
L_{cls}(p, t) = -(t\log(p) + (1-t)log(1-p)).
\end{split}
\end{equation}
In our experiments, we use $\gamma=10$. Due to the imbalanced distributions of positive and negative samples, we normalize their loss separately. OHEM~\cite{ohem} is applied to the negative term of the classification loss. Each anchor is parameterized as ($x_a, y_a, z_a, h_a, w_a, l_a, \theta_a)$ and the ground truth box is parameterized as ($x_g, y_g, z_g, h_g, w_g, l_g, \theta_g$). For regression, we adopt parameterization following \cite{VoxelNet, fastrcnn} as
\begin{equation}\label{equ:voxelrpn_loss}
\begin{split}
\Delta_{1} x &= \frac{x_g - x_a}{d_a}, \Delta_{1} y = \frac{y_g - y_a}{d_a}, \Delta_{1} z = \frac{z_g - z_a}{h_a}, \\
\Delta_{1} h &= \log(\frac{h_g}{h_a}), \Delta_{1} w = \log( \frac{w_g}{w_a}), \Delta_{1} l = \log( \frac{l_g}{l_a}),\\
\Delta_{1} \theta &= \theta_g - \theta_a.\\
\end{split}
\end{equation}
The regression loss is defined as a smooth L1 loss of
\begin{equation}
  L_{reg}(x)=\left\{
  \begin{array}{ll}
    0.5(\sigma x)^2, & \text{if}\ |x| < 1/\sigma^2 \\
    |x| - 0.5/\sigma^2, & \text{otherwise}
  \end{array}\right.
\end{equation}
where $\sigma$ is set to $3$ in our experiments.

\vspace{-0.1in}
\paragraph{RefinerNet} It is noticed that the recall of our VoxelRPN on 0.5 IoU thresh, in top 30 predicted boxes in Bird's Eve View (BEV), is over 95\% for car. Our RefinerNet is for improving the quality of prediction boxes. We only train it on positive proposal boxes whose IoU with ground-truth is higher than 0.5 in BEV. 

The regression target is defined as the offset from proposal center $(x_p, y_p, z_p)$ to 8 canonized corners ($x_{i,g}, y_{i,g}, z_{i,g}$ for $i = 1,...,8$) of the target box as shown in Figure~\ref{fig:proposal}:
\begin{equation}\label{equ:pointrcnn_loss}
\begin{split}
\Delta_{2} x_i = x_{i, g} - x_{p}, \Delta_{2} y_i = y_{i, g} - y_{p}, \Delta_{2} z_i = z_{i,g} - z_{p} 
\end{split}
\end{equation}
This parameterization is a general and natural design for RefinerNet that processes directly on coordinates of points. 



\section{Experiments}
We conduct experiments on the challenging KITTI \cite{kitti} dataset in terms of 3D detection and BEV detection. Extensive ablation studies on our approach are conducted.

\subsection{Experiment Setup}
\paragraph{Dataset and Evaluation Metric}
    The KITTI dataset provides 7,481 images and point clouds for training and 7,518 for testing. Note for evaluation on the test subset and comparison with other methods, we can only submit our result to the evaluation server. Following the protocol in~\cite{MV3D, VoxelNet}, we divide the training data into a training set (3,712 images and point clouds) with around 14,000 Car annotations and a validation set (3,769 images and point clouds). Ablation studies are conducted on this split. While for evaluation on test set, we train our model on the entire train set with 7k point clouds.

    According to the occlusion/truncation level and the height of 2D boxes in images, evaluation on the KITTI dataset is split into three difficulty levels as ``easy", ``moderate" and ``hard". The KITTI leaderboard ranks all methods according to $\text{AP}_{0.7}$ in ``moderate" difficulty and takes it as the primary metric. 

\vspace{-0.1in}
\paragraph{Implementation Details}
    The point cloud is cropped to the range of $[0., 70.4] \times [-40., 40.] \times [-3., 1.]$ meters along $(X, Y, Z)$ axes respectively, following \cite{MV3D, VoxelNet}. The input to VoxelRPN is generated by voxelizing the point cloud into a 3D cuboid of size $800 \times 704 \times 20$, where each voxel is with size $0.1 \times 0.1 \times 0.2$ meter. As a result, the output convolutional feature map is with size $ 200\times 176\times 1$. 4 anchors are defined in each output location with different angles ($0\degree, 45\degree, 90\degree, 135\degree$). 
    
    For the category of ``car", we use the anchor size of $h_a = 1.73, w_a=0.6, l_a=0.8$ meters. NMS with IoU threshold 0.1 is applied to prediction from VoxelRPN to filter out duplicated predictions and help keep high efficiency of the RefinerNet. For the categories of \textit{Pedestrian} and \textit{Cyclist}, the network removes the downsampling in the fourth Conv3D layer since these two categories are much smaller than car category. 
    
    We use anchors of size $h_a = 1.73, w_a=0.6, l_a=0.8$ and $h_a = 1.73, w_a = 0.6, l_a=1.76$ for \textit{Pedestrian} and \textit{Cyclist} respectively. Like F-PointNet\cite{fpointnet}, multi-class prediction for RefinerNet is to concatenate predicted class label of VoxelRPN (one-hot encoding vector) with the feature after max-pooling operation and then refine box corners for all classes. We note that training on \textit{Pedestrian} and \textit{Cyclist} can improve their performance.

\vspace{-0.1in}
\paragraph{Training Details}
    By default, models are trained on 8 NVIDIA P40 GPUs with batch-size 16 -- that is, each GPU holds 2 point clouds. We apply ADAM \cite{adam} optimizer with an initial learning rate $0.01$ for training of VoxelRPN and RefinerNet. We train VoxelRPN for 70 epochs and the learning rate is decreased by 10 times at 50th and 65th epochs. Training of RefinerNet lasts for 70 epochs and the learning rate is decreased by 10 times at 40th, 55th and 65th epochs. 
    
    Batch Normalization is used following each parameter layer. A weight decay of $0.0001$ is used in both networks. Since the training of RefinerNet requires the convolutional feature from VoxelRPN, we train it for each frame instead of on objects, saving a large amount of computation. 
    
\vspace{-0.1in}
\paragraph{Data Augmentation}
	Multiple data augmentation strategies are applied during training in order to alleviate the overfitting problem considering the limited amount of training data. For each frame of the point cloud, we conduct left-right random flipping, random scaling with a uniformly sampled scale from $0.95 \sim 1.05$ and random rotation with a degree sampled from $-45\degree \sim 45\degree$ around the origin for entire scene of point clouds. 
	
	We also disturb each ground-truth bounding box and its corresponding interior points by random translation. Specifically, the shift is sampled from $\mathcal{N}(0, 1)$ for both $X$ and $Y$ axes and $\mathcal{N}(0, 0.3)$ for $Z$ axis. Random rotation around $Z$ axis is uniformly sampled from $-18\degree \sim 18\degree$. Note that there is a collision detection to prevent collision of different objects. 

\vspace{-0.1in}
\paragraph{MIXUP Augmentation}
Similar to the spirit of \cite{cut&paste,mixup} in 2D object detection, we also augment input point clouds with cropped ground-truth from other point sets to greatly improve the convergence speed and quality. 
Instead of cropping solely interior points of each ground-truth box, we crop a larger region with extra 0.3 meters to better preserve the context information. With this regularization, cropped points and surrounding points are distributed more coherently with each other, making the network better capture the property of each object. In our setting, 20 objects are added in each frame of point clouds.


\subsection{Main Results}

\begin{table*}[bpt]\footnotesize
\begin{center}
\begin{tabular}{l|c|c|c|c|c|c|c|c|c}
\multirow{2}{*}{Method} & \multirow{2}{*}{Input} & \multirow{2}{*}{Time (s)} & \multicolumn{3}{c|}{3D} & \multicolumn{3}{c|}{BEV} & \multirow{2}{*}{GPU}\\ \cline{4-9}
& & & $\text{AP}_{easy}$ & {\bf $\text{AP}_{\mathbf{moderate}}$} & $\text{AP}_{hard}$ & $\text{AP}_{easy}$ & {\bf $\text{AP}_{\mathbf{moderate}}$} & $\text{AP}_{hard}$ \\ \hline \hline
MV3D~\cite{MV3D} & L+I & 0.24 & 66.77 & 52.73 & 51.31 & 85.82 &  77.00 & 68.94 & TITAN X \\ \hline
AVOD-FPN~\cite{AVOD} & L+I & 0.1 & 81.94 & 71.88 &  66.38 &  88.53 &  83.79 &  {\bf 77.90} & TITAN XP \\ \hline
AVOD~\cite{AVOD} & L+I & 0.1 & 73.59 & 65.78 & 58.38 & 86.80 &  85.44 & 77.73 & TITAN XP \\ \hline
F-PointNet~\cite{fpointnet} & L+I & 0.17 & 81.20 & 70.39 & 62.19 & 88.70 & 84.00 & 75.33 & GTX 1080 \\ \hline
ContFuse~\cite{contfuse} & L+I & 0.06 & 82.54 & 66.22 & {\bf 64.04} & {\bf 88.81} & {\bf 85.83} & 77.33 & -- \\ \hline
RoarNet~\cite{roarnet} & L+I & 0.1 & {\bf 83.71} & {\bf 73.04} & 59.16 & 88.20 & 79.41 & 70.02 & TITAN X\\ \hline 
IPOD~\cite{ipod} & L+I & 0.2 & 79.75 & 72.57 &	66.33 & 86.93 &	83.98 &	77.85 & Tesla P40 \\ \hline
\hline
VoxelNet~\cite{VoxelNet} & L & 0.22 & 77.49 & 65.11 & 57.73 & {\bf 89.35} & 79.26 & 77.39 & TITAN X\\ \hline
PIXOR~\cite{PIXOR} & L & 0.1 & - & - & - & 84.44 & 80.04 & 74.31 & TITAN XP \\\hline
SECOND~\cite{second} & L & 0.05 & 83.13 & 73.66 & 66.20 & 88.07 & 79.37 & 77.95 & GTX 1080Ti\\ \hline
PointPillars~\cite{pointpillar} & L & 0.016 & 79.05 &  74.99 & {\bf 68.30} & 88.35 &	{\bf 86.10} & {\bf 79.83} & GTX 1080Ti \\ \hline
PointRCNN-deprecate~\cite{pointrcnn} & L & 0.1 & 84.32 & 75.42 & 67.86 & 89.28 & 86.04 & 79.02 &  TITAN XP\\ \hline
PointRCNN~\cite{pointrcnn} & L & 0.1 & {\bf 85.94} & \bf 75.76 & \bf 68.32 & \bf 89.47  & 85.68  & \bf 79.10 & TITAN XP \\ \hline
Fast Point R-CNN & L & 0.065 & 84.28 & 75.73 & 67.39 & 88.03 & {\bf 86.10} & 78.17 & Tesla P40  \\
\end{tabular}
\end{center}
\caption{Comparison of main results on KITTI test set. Here `L' denotes LiDAR input and `I' denotes RGB image input.}
\label{table: Comparison of main results on KITTI test set.}
\end{table*}

\begin{table*}[bpt]\footnotesize
\begin{center}
\begin{tabular}{l|c|c|c|c|c|c|c}
\multirow{2}{*}{Method} & \multirow{2}{*}{Time (s)} & \multicolumn{3}{c|}{3D} & \multicolumn{3}{c}{BEV } \\ \cline{3-8}
& & $\text{AP}_{easy}$ & {\bf $\text{AP}_{\mathbf{moderate}}$} & $\text{AP}_{hard}$ & $\text{AP}_{easy}$ & {\bf $\text{AP}_{\mathbf{moderate}}$} & $\text{AP}_{hard}$ \\
\hline\hline
VoxelNet (Paper) & 0.225 & 81.97 & 65.46 & 62.85 & 89.60 & 84.81 & 78.57 \\
\hline
VoxelNet (Reproduced) & 0.117 & 86.48 & 75.26 & 73.25 & 90.13 & 87.61 & 86.4 \\
\hline
VoxelRPN & 0.058 & 87.51 & 76.64 & 74.4 & 89.8 & 87.58 & 86.38 \\
\hline
Fast Point R-CNN & {\bf 0.065} & {\bf 89.12} & {\bf 79.00} & {\bf 77.48} & {\bf 90.12} & {\bf 88.10} & {\bf 86.24} \\
\end{tabular}
\end{center}
\caption{Comparison of main results on KITTI validation set. }
\label{table:Comparison of main results on KITTI validation set.}
\end{table*}

As shown in Table~\ref{table: Comparison of main results on KITTI test set.}, we compare Fast Point R-CNN with state-of-the-art approaches in 3D object detection and BEV object detection on KITTI test dataset. The official KITTI benchmark ranks different methods according to the performance on the moderate subset. Our model achieves state-of-the-art performance while accomplishing high efficiency (15FPS on NVIDIA Tesla P40 GPU). Note that SECOND~\cite{second} applies SparseConv~\cite{sparseconv} and PointPillars~\cite{pointpillar} used engineering techniques of NVIDIA TensorRT. These solutions are complementary to ours.

\begin{figure*}
    \begin{center}
        \includegraphics[width=1\linewidth]{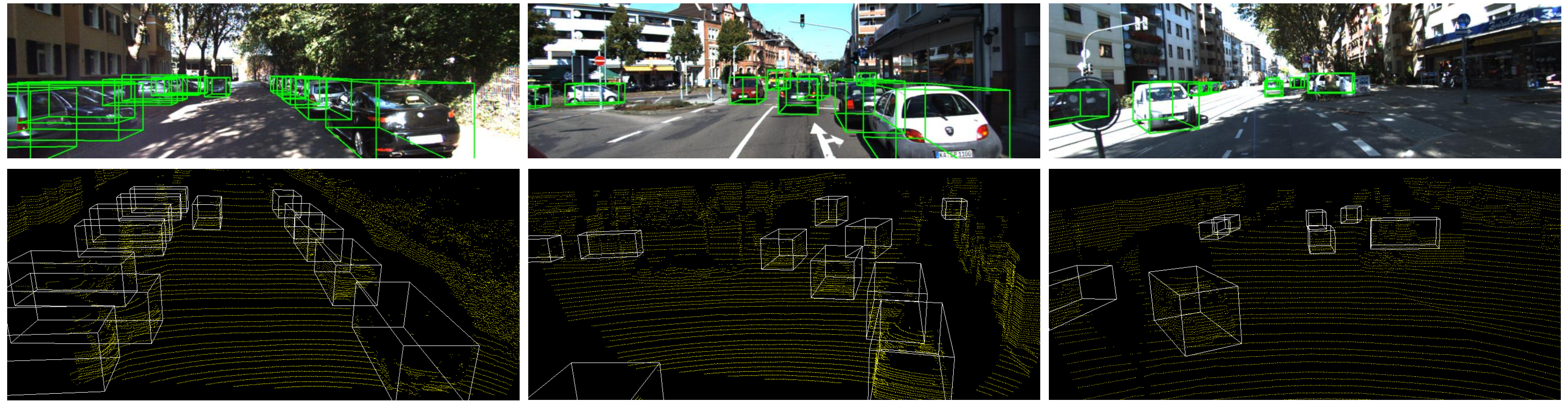}\vspace{-0.1in}
    \end{center}
    \caption{Visualization of our results.}
    \label{fig:results}
\end{figure*}

For better comparison, we reproduce VoxelNet~\cite{VoxelNet} as a strong baseline network. It is noteworthy that our reproduction even yields much better results than those reported in \cite{VoxelNet}. As shown in Table~\ref{table:Comparison of main results on KITTI validation set.}, our proposed VoxelRPN outperforms VoxelNet in 3D object detection. Accompanied by RefinerNet, nearly twice as fast as VoxelNet, Fast Point R-CNN outperforms VoxelNet in both 3D object detection and BEV object detection. We show qualitative results in Figure~\ref{fig:results}. We can make good prediction at several challenging scenes.

\section{Ablation Studies}
We conduct extensive ablation study for each component based on the train/val. split.

\subsection{VoxelRPN}
To illustrate the effectiveness of VoxelRPN, we start with a fast and yet simple baseline and gradually add our proposed components. The baseline consists of only 2D convolutions and directly processes input voxel by encoding information along $Z$ axis into the channel dimension. The difference with VoxelRPN is that the first 6 Conv3D layers in the first block are replaced with 6 Conv2D layers. We keep the same kernel size in $X$ and $Y$ axes; the channels are 128 except the first layer with 64 channels. Two anchors with angles $0\degree$ and $90\degree$ are used. As shown in Table~\ref{table:voxelrpn techniques}, the baseline achieves reasonable performance.

\vspace{-0.1in}
\paragraph{More 3D Convolutions (Conv3D)}  By replacing lower layers to 3D convolutions as illustrated in Figure~\ref{fig:voxelrpn} and processing the 3D voxels, we improve the baseline by nearly $1$ point, manifesting the effectiveness of 3D convolutions on preserving the information, especially along $Z$ dimension. With this modification, the time cost only increases 5ms.

\vspace{-0.1in}
\paragraph{Higher Resolution Input (HRI)} We also introduce the finer voxel, producing higher resolution grid input with size $800 \times 704 \times 20$, as described in Figure~\ref{fig:voxelrpn}. Accordingly, we modify the stride of the first layer to 2 to effectively reduce the computation overhead. This technique can significantly improve the results without adding much computation.

\vspace{-0.1in}
\paragraph{MIXUP Augmentation (MIXUP)} With MIXUP augmentation, we improve the performance with around 0.5 point. With MIXUP augmentation, we achieve comparable performance with only half of the original training epochs.

\vspace{-0.1in}
\paragraph{More Anchors (MA)} With $4$ anchors in angles $0\degree$, $45\degree$, $90\degree$ and $135\degree$ respectively, instead of using only 2 anchors, we further gain another $0.8$ point bonus. We find that the matching probability gain with ground-truth is significant with more anchors involved.

\begin{table}[bpt]\small
    \begin{center}
        \begin{tabular}{c|c|c|c|c}
        Conv3D & HRI & MIXUP & MA & 3D $\text{AP}_{0.7}$ (moderate)\\ \hline
        \hline
        - & - & - & - &  73.8 \\ \hline
         \checkmark & & & & 74.7 \\ \hline
         \checkmark & \checkmark & & & 75.34 \\ \hline
         \checkmark & \checkmark & \checkmark & & 75.82 \\ \hline
         \checkmark & \checkmark & \checkmark & \checkmark & 76.64 \\
        \end{tabular}
    \end{center}
    \caption{Effectiveness of different techniques applied to VoxelRPN on KITTI val subset. The baseline network consists of only 2D convolutions. Conv3D denotes we replace the lower layers with 3D convolutions. HRI denotes high resolution input. MIXUP denotes the use of MIXUP augmentation. MA means anchors with 4 different angles are used instead of 2 of them.}
    \label{table:voxelrpn techniques}
\end{table}

\subsection{RefinerNet}
\paragraph{Input Features}
We first investigate the importance of both coordinate and convolution features. As shown in Table~\ref{table:comparison of different fusion methods}, with only coordinate feature or convolution feature, the RefinerNet improves results over VoxelRPN. It is noticeable that the performance with coordinate feature as input is better than the one with convolution feature as input. This manifests that the accurate location information is lost in the quantization representation of point cloud and consecutive convolutional-and-downsampling operations. 

\vspace{-0.1in}
\paragraph{Feature Fusion}
With the compensation of coordinate information, the performance boosts greatly. Much better performance is achieved with both coordinate and convolution features, since they provide semantically complementary information. 
We also compare our strategy of fusing these two sources of features with simple concatenation. Our fusion method with attention mechanism outperforms the alternative by 0.62 point, as shown in Table~\ref{table:comparison of different fusion methods}. 

\begin{table}[bpt]\small
    \begin{center}
        \begin{tabular}{l|c}
        Fuse methods & 3D $\text{AP}_{0.7}$ (moderate) \\ \hline \hline
        Coordinate Feature & 77.82 \\ \hline
        Convolution Feature & 76.90 \\ \hline
        Concatenation & 78.38 \\ \hline
        + Attention Module & 79.00 \\ \hline
        \end{tabular}
    \end{center}
    \caption{Comparison of different fusion methods in RefinerNet. }
    \label{table:comparison of different fusion methods}
\end{table}

\vspace{-0.1in}
\paragraph{Effect of Canonized Corner Loss}
We compare parameterization of box prediction. The naive parameterization of 7 parameters as regression loss only achieves 78.45 in 3D $\text{AP}_{0.7}$. With canonized corner loss, it can further improve to 79.

\vspace{-0.1in}
\paragraph{Comparison with RoI Align}
One straightforward method for box refinement is to use RoI Align\cite{maskrcnn}. For comparison, we implement rotated RoI align that crops convolutional features from VoxelRPN given proposals. For the car class, we pool with size $8\times 4$ along the direction of car inside the rotated box region. Then two 4096D MLP layers are applied to perform classification and regression. Only the above operations are different -- it achieves 77.39 with $\text{AP}_{0.7}$. Our RefinerNet performs better clearly. We conjecture that rotated RoI align still lacks precise localization information.

\vspace{-0.1in}
\paragraph{Result Analysis}

\begin{table}[bpt]\footnotesize
\begin{center}\addtolength{\tabcolsep}{-3pt}
\begin{tabular}{l|c|c|c|c|c}
\multirow{2}{*}{Method} & \multirow{2}{*}{Range (meters)} & \multicolumn{2}{c|}{3D (Moderate)} & \multicolumn{2}{c}{BEV (Moderate)}  \\ \cline{3-6}
& & $\text{AP}_{0.7}$ & $\text{AP}_{0.8}$ & $\text{AP}_{0.7}$ & $\text{AP}_{0.8}$ \\
\hline\hline
VoxelRPN & 0-30 & 88.39 & 58.81 & 90.22 & 83.32 \\
\hline
Fast Point R-CNN & 0-30 & 89.26 & 62.73 & 90.25 & 85.61 \\
\hline \hline
VoxelRPN & 30-50 & 51.99 & 13.31 & 73.51 & 49.63  \\
\hline
Fast Point R-CNN & 30-50 & 58.41 & 15.39 & 73.9 & 50.05 \\
\end{tabular}
\end{center}
\caption{Comparison of nearby- and distant-object detection accuracy.}
\label{table:comparison of near and far objects}
\end{table}

In the scenario of autonomous driving, faraway objects are with much less points due to the limited resolution of LiDAR and occlusion by nearby objects, making it more challenging to detect distant objects. As shown in Table~\ref{table:comparison of near and far objects}, there is a large discrepancy between accuracy of nearby and faraway objects. It is noteworthy that RefinerNet significantly improves the performance of 3D detection accuracy of distant objects ranging from $30$ to $50$ meters, \ie, from $51.99$ to $58.41$ with $\text{AP}_{0.7}$ metric. It is because distant objects generally possess only a small number of points. With only voxel representation, it is hard for VoxelRPN to fully capture the structure of objects. But with the profitable access to coordinate feature, RefinerNet can still infer the complete structure of objects and achieve better inference.

As shown in Tables~\ref{table:comparison of near and far objects} and \ref{table:comparison of different AP thresh}, RefinerNet can further improve detection with higher quality, evaluated with $\text{AP}_{0.8}$, which demonstrates that RefinerNet better utilizes fine-grained localization information than VoxelRPN.

\begin{table}\footnotesize
\begin{center}\addtolength{\tabcolsep}{-2pt}
\begin{tabular}{l|c|c|c|c|c|c}
\multirow{2}{*}{Method} & \multicolumn{3}{c|}{3D (Moderate)} & \multicolumn{3}{c}{BEV (Moderate)}  \\ \cline{2-7}
& $\text{AP}_{0.6}$ & $\text{AP}_{0.7}$ & $\text{AP}_{0.8}$ & $\text{AP}_{0.6}$ & $\text{AP}_{0.7}$ & $\text{AP}_{0.8}$ \\
\hline\hline
VoxelRPN & 88.94 & 76.64 & 42.6 & 89.77 & 87.58 & 71.39 \\
\hline
Fast Point R-CNN & 89.14 & 79.0 & 52.95 & 89.86 & 88.10 & 74.58 \\
\end{tabular}
\end{center}
\caption{Detection results with different IoU thresholds.}
\label{table:comparison of different AP thresh}
\end{table}

\subsection{Experiments on Other Categories}
KITTI benchmark provides limited annotations for \textit{Pedestrian} and \textit{Cyclist} categories. For reference, we provide results on these two classes. Following \cite{VoxelNet, second}, we train the network for these two categories. Our final results on \textit{Pedestrian} and \textit{Cyclist} are 63.05 and 64.32 respectively, with VoxelRPN results 60.78 and 62.41 on KITTI val dataset. We achieve comparable results on KITTI test data as listed in Table~\ref{table:experiments on other classes}. We believe when more data is used, superiority of our two-stage network can be better demonstrated.


\begin{table}[t]\footnotesize
	\begin{center}\addtolength{\tabcolsep}{-2pt}
		\begin{tabular}{l|c|c|c|c}
			
			\multirow{2}{*}{Method} & \multicolumn{2}{c|}{$\text{AP}_{0.5}$ on \textit{Pedestrian}} & \multicolumn{2}{c}{$\text{AP}_{0.5}$ on \textit{Cyclist}}  \\ \cline{2-5}
			& 3D  & BEV  & 3D  & BEV  \\ \hline\hline
			PointPillars~\cite{pointpillar} & 43.53 & {\bf 50.23} & 59.07 & 62.25 \\ \hline
			F-PointNet~\cite{fpointnet} & {\bf 44.89} & 50.22 & 56.77 & 61.96 \\ \hline 			
			PointRCNN~\cite{pointrcnn} & 41.78 & -- & \bf 59.60 & -- \\ \hline
			Fast Point R-CNN & 42.90 & 45.43 & 59.36 & {\bf 62.59} \\ \hline
		\end{tabular}
	\end{center}\vspace{-3mm}
	\caption{Performance on \textit{Pedestrian} and \textit{Cyclist} on test set.}
	\label{table:experiments on other classes}
\end{table}

\section{Conclusion}
In this paper, we have proposed a generic, effective and fast two-stage framework for 3D object detection. Our method makes use of both voxel representation and raw point cloud to benefit from both of them. The first stage takes voxel representation as input and applies convolutional operations to acquire a set of initial predictions. Then the second stage further refines them based on raw point clouds and extracted convolution features. 

With this conceptually simple but practically powerful design, our method is on par with existing solutions while maintaining higher detection speed. We believe our research shows a new way to properly utilize different dimensions of information for this challenging and yet practically fundamental task.

{\small

}

\end{document}